\documentclass[runningheads]{llncs}
\usepackage{graphicx}
\newcommand{\repeatthanks}{\textsuperscript{\thefootnote}}
\begin{document}
%
\title{Comparative Study of Long Document Classification}

%
%
\author{Vedangi Wagh\inst{1}\thanks{Authors contributed equally} \and
Snehal Khandve\inst{1}\repeatthanks \and
Isha Joshi\inst{1}\repeatthanks\and
Apurva Wani\inst{1}\repeatthanks\and
Geetanjali Kale\inst{1}\and
Raviraj Joshi\inst{2}
}

\authorrunning{Vedangi Wagh et al.}

%
\institute{Pune Institute of Computer Technology, Pune \and
Indian Institute of Technology Madras, Chennai\\
\email{{\{vedangikwagh, snehal.khandve07, ishajoshi.211, apurva.wani06, ravirajoshi\}@gmail.com\\
gvkale@pict.edu}}}
%
\maketitle              
\begin{abstract}
The amount of information stored in the form of documents on the internet has been increasing rapidly. Thus it has become a necessity to organize and maintain these documents in an optimum manner. Text classification algorithms study the complex relationships between words in a text and try to interpret the semantics of the document. These algorithms have evolved significantly in the past few years. There has been a lot of progress from simple machine learning algorithms to transformer-based architectures. However, existing literature has analyzed different approaches on different data sets thus making it difficult to compare the performance of machine learning algorithms. In this work, we revisit long document classification using standard machine learning approaches. We benchmark approaches ranging from simple Naive Bayes to complex BERT on six standard text classification datasets. We present an exhaustive comparison of different algorithms on a range of long document datasets. 
We re-iterate that long document classification is a simpler task and even basic algorithms perform competitively with BERT-based approaches on most of the datasets. The BERT-based models perform consistently well on all the datasets and can be blindly used for the document classification task when the computations cost is not a concern. 
In the shallow model's category, we suggest the usage of raw BiLSTM + Max architecture which performs decently across all the datasets. 
Even simpler Glove + Attention bag of words model can be utilized for simpler use cases. The importance of using sophisticated models is clearly visible in the IMDB sentiment dataset which is a comparatively harder task.

\end{abstract}
\section{Introduction}
The text has been generated at an unprecedented pace for the past few decades. One of the major reasons for this is the digitization of information which is a cheaper and faster means of transfer than conventional methods. As more and more information is stored and shared in the form of text files, it becomes difficult to analyze such large quantities of text data. Natural Language Processing (NLP) tasks delve into the processing and analysis of text to extract information and insights. It consists of a wide range of tasks like text classification, natural language inference, machine translation, question answering, etc \cite{wolf2020transformers}. All these tasks are aimed at solving a particular problem and providing a better interpretation of natural language data.

Text classification is a problem designed to assign a piece of text a proper class or category \cite{zhang2015character,conneau2016very}. The classes depend on the domain under consideration and range from a variety of topics. The text classification techniques can be used to categorize, store and maintain different kinds of text available on the internet like - medical records, documents, research papers, product reviews, movie reviews, etc. Hence, text classification has many applications like spam detection, sentiment analysis, document classification, search engines, topic labeling, hostility detection.

This paper particularly emphasizes long document classification \cite{kowsari2019text}\cite{pappagari2019hierarchical} as opposed to short text classification \cite{lee2016sequential}. A long document consists of multiple paragraphs which in turn have many sentences. Thus, in long documents, we have numerous words that impact the meaning or class of the document. As the document size becomes larger it becomes difficult to maintain the context over a large length. The task might also become simpler due to the large availability of important signals necessary for classification.
We analyze documents of varying characteristics and their performance on various classification algorithms.

We have studied the recent developments in deep learning techniques for text classification. In these methods text at the word or sentence level is represented in the form of a fixed-size vector called embeddings. These embeddings encode semantic and contextual information about the word or sentence which is then processed by the model to interpret the text. Specifically, deep learning techniques models the complex relationships between these embeddings to assign it the best possible category \cite{taware2021shuftext,minaee2021deep}.

Text classification models have advanced from vanilla models like Naive Bayes to complex models like BERT. In this work, we consider models based on Naive Bayes, CNN, LSTM, and pre-trained transformer-based architectures. The pre-trained architectures used are USE, ULMFiT, BERT, and DistilBERT. Shallow neural network architectures like LSTM+CNN, BiLSTM, HAN are trained from scratch. Along with these deep learning approaches, a non-neural approach like the Naive Bayes algorithm using TFIDF features is also evaluated for comparison. The datasets used are mostly a collection of news articles, sports-based articles, and sentiment datasets. In total 6 datasets are selected for this paper which are BBC News, AG News, 20 NewsGroup (20 NG), R8, BBC Sports, and IMDB.

Recent works in long document classification have focused on optimizing transformer models. The quadratic dependency of memory and time on sequence length is the major drawback of transformer architecture. Modifications like Longformer \cite{beltagy2020longformer}, Linformer \cite{wang2020linformer}, Reformer \cite{kitaev2020reformer}, and BigBird \cite{zaheer2020big} have been proposed in literature to process long documents. Similarly, adaptations of BERT for long document classification have been explored in \cite{adhikari2019docbert,ding2020cogltx}. In this work, we are not specifically concerned with the efficiency of transformer models and instead, look at different families of architectures. Base-BERT and DistilBERT from the transformer family are explored in this work. Alternatively, hierarchical approaches have been popular for long document classification \cite{pappagari2019hierarchical} \cite{yang2016hierarchical}. We have used Hierarchical Attention Networks (HAN) from this family. The related works in literature have considered different datasets and algorithms thus making it difficult to have a holistic view \cite{li2020survey}. In this work, we provide a comparative view of different families of algorithms on a range of datasets. Similar comparison of deep learning approaches on different datasets and languages have been studied in \cite{kamath2018comparative,joshi2019deep,elnagar2020arabic,zulqarnain2020comparative,gonzalez2020comparing,kulkarni2021experimental}.

\section{Dataset Details}
In Table \ref{tab1} statistics of the dataset are reported without any preprocessing. The entire dataset is used to calculate the unique tokens and the average length of the documents.
\begin{itemize}

\item BBC News:
This dataset \cite{greene06icml} contains a total of 2225 records. Using a 20\% split, test data contains 445 records and the remaining 1780 records are in the train data. These records are News articles which are classified into 5 categories. The categories are Business, Politics, Sport, Entertainment, and Tech.
\item AG News:
The dataset \cite{AGNews:online} is a collection of 127600 records. There are 120000 records in the training set while there are 7600 records in the testing set. Each record is a news article that falls into either of the 4 categories viz., World, Sports, Business, and Sci/Tech. These articles were collected by ComeToMyHead from more than 2000 news sources.
\item 20NG:
In the 20 Newsgroups dataset\cite{scikit-learn}, each record is text taken from 20 different newsgroups. This dataset is popularly used for text classification and text clustering tasks. There are a total of 18774 records.
\item BBC Sports:
This dataset \cite{greene06icml} consists of sports articles gathered from the BBC Sport website. It has five classes to classify these articles into viz., athletics, cricket, football, rugby, and tennis. It has a total of 737 articles. Of these 147 articles are in the test data and the rest 590 articles are in the train data.
\item IMDB:
This dataset \cite{maas-EtAl:2011:ACL-HLT2011} is a collection of movie reviews and is useful for binary sentiment classification. The two classes are either positive or negative. It has a total of 50000 reviews, 25000 for training, and 25000 for testing the model. 
\item R8:
This dataset \cite{R8:online} is a part of Reuters 21578 datasets. The dataset was gathered and labeled by Carnegie Group, and Reuters Ltd during the development of the CONSTRUE text categorization system. Headquartered in London. Reuters, Ltd. is an international news agency.  The documents are divided into 8 classes. These 8 classes are acq, earn, crude, grain, money-fx, ship, interest, and trade. The train dataset has 5485 records while the test dataset has 2189 records.
\end{itemize}

\begin{table}
 \caption{Summary statistics of datasets}
 \label{tab1}
 \begin{center}
\begin{tabular*}{\textwidth}{c @{\extracolsep{\fill}} cccccc}
 \hline
 Dataset & Docs & Training & Test & Classes & Unique & Avg Length \\ & & & & & Tokens &  (words)\\
 \hline
BBC News & 2225 & 1780 & 445 & 5 & 32360 & 389 \\
\hline
AG News & 127600 & 120000 & 7600  & 4 & 72046 & 39 \\
\hline
20 NG & 18846 & 11314 & 7532 & 20 & 179209 & 315  \\
\hline
BBC Sports & 737 & 590 & 147 & 5 & 14224 & 337 \\
\hline
IMDB & 50000 & 25000 & 25000 & 2 & 124252 & 231 \\
\hline
R8 & 7674 & 5485 & 2189 & 8 & 16698 & 64 \\
\hline
 \end{tabular*}
 \end{center}
\end{table}

\section{Experimental Setup}
All of the models are trained using the Tensorflow 2.0 framework with batch size ranging from 16 to 128 depending on the dataset. Throughout the experiments, the Adam optimizer was employed with gradient descent and cross-entropy loss function. A highest of 15 epochs was used and the optimal epoch was chosen using a validation set derived from train data.
The pre-processing steps were the same as the ones used in \cite{wani2021evaluating}.

\subsection{Experiments}
In this section, the various categorization strategies that were employed on the above-mentioned datasets are described.
\begin{itemize}
    \item ULMFiT - This is an architecture presented in \cite{howard2018universal} which uses the approach of fine-tuning a language model for a specific task. In this approach, the general domain language model training is only done once and then the model is fine-tuned specifically to the target task. For the implementation of ULMFit, the fastai library is used. The publicly available pre-trained model is fine-tuned for all the datasets.
    \item USE - The Universal Sentence Encoder architecture described in \cite{cer2018universal} is an encoder that converts text or sentences or words in a 512-dimensional vector. This pre-trained encoder publicly available on TensorFlow-hub can be used for various natural language tasks like text classification. The entire text is encoded into a high-dimensional vector and passed as an input embedding to the network which consists of two sets of dense, dropout, and batch normalization layers in succession for the purpose of fine-tuning the model to the specific dataset. Dense layers of sizes 256 and 128 are used. A dropout rate of 0.4 is employed.
    \item BiLSTM + Max - This is based on one of the architectures mentioned in this \cite{conneau2017supervised}, we have used a single BiLSTM layer followed by max-pooling architecture. 100-dimensional GLoVe embeddings are passed on to a Bidirectional LSTM layer with 256 units. The Bi-LSTM network processes the input sequence to generate a sequence of 256 vectors which are then passed on to a max-pooling layer. This layer outputs a vector with a maximum value of 256 vectors over each dimension. This layer is further connected to a dropout layer with a rate of 0.5 and 1 dense layer with 128 units.
    \item BiLSTM + Attention - This is similar to previous architecture with max-pooling being replaced by an attention mechanism. The structure of the attention layer is was inspired by \cite{zhou2016attention}. This model has a first embedding layer of 100-dimension GLoVe vectors which is then connected to 2 Bi-LSTM layers with 256 and 128 units respectively. These layers are followed by an attention layer and a dense layer of 128 units.
    \item LSTM + CNN - This architecture uses LSTM and CNN layers in succession followed by max pooling. 100-dimensional GLoVe word embeddings of the given text were provided as input to the network. This network consisted of an LSTM layer with 128 units connected to a CNN layer with 250 filters of size 2. This is followed by max-pooling and a dense layer of 256 units to fine-tune the model. 
    \item Transformers - Transformers have surpassed previous sequential models in performance in almost all-natural language processing tasks. A major advantage of transformers over RNNs is that it processes the input parallelly which leads to maximum utilization of the contemporary hardware. An important element of transformers is self-attention which is used for generating a contextual embedding of a word concerning other words in the input text.
    We have experimented with two types of transformer-based architectures on the datasets. 
    \begin{itemize}
        \item BERT - The BERT-base architecture comprises 12 transformer blocks and uses a hidden size of 768. It uses self-attention with 12 attention heads. The model receives 512-word embeddings as input and creates a representation for the input sequence. For dividing segments, a special token [SEP] is used, which is useful for some NLP tasks. For text categorization, however, [CLS] embedding, which is always the first token in the sequence, is used. To categorize the representation, the embedding of the [CLS] token taken from the last layer is passed through a softmax classifier. To implement the BERT architecture the pre-trained ‘bert-base-uncased' model from the simpletransformers \cite{wolf2020transformers} library has been utilized.
        \item DistilBERT - DistilBERT is a variant of BERT with the same basic transformer architecture. It offers a simpler, lighter, and cheaper alternative using the process of distillation. Distillation is carried out at the process's pre-training phase. The number of layers is halved and the algebraic operations are optimized for this aim. Using these techniques, even though DistilBERT is 40 \% smaller than BERT, it provides competitive results. Similar to the BERT architecture the pre-trained ‘distilbert-base-uncased' model from the simpletransformers library has been utilized for the implementation of distilbert architecture.
    \end{itemize}
    
    \item TFIDF using Naive Bayes - TFIDF is a weighted word technique of feature extraction which is used to reflect how important a word is to a document in a corpus. TF which stands for term frequency simply measures the frequency at which a word occurs in the document. However, there are some commonly used words that do not contribute semantically to the document. Hence IDF (Inverse Document Frequency) is used along with TF to reduce the effect of commonly used words. It assigns more weightage to the words with very high or very low frequency. The word vectors generated for a particular document using TF-IDF are passed to a Naive Bayes classifier.
    
    \item GloVe + Average - In this model, a 100-dimensional GloVe embedding layer is followed by a global average pooling layer and a dense layer of 128 units.
    
    \item GloVe + Attention - In this model, a 100-dimensional GloVe embedding layer is followed by an attention layer and a dense layer of 128 units. The attention layer is described in \cite{zhou2016attention}.
    
    \item HAN - Hierarchical Attention Network(HAN) is a type of LSTM-based architecture that consists of two levels of the encoder at word level and sentence level. The word encoder is followed by word-level attention and the subsequent sentence encoder is followed by sentence-level attention. The input sequences are split into no more than 40 sentences, with each phrase containing no more than 50 words. The word encoder, which is a bidirectional LSTM, creates representations for individual words in the first phase. This information is passed on to the word-level attention layer, which recognizes keywords in the sentence that contribute to its meaning. A sentence vector is created by merging essential words that define the phrase's meaning. The sentence encoder which is a Bidirectional LSTM process this sentence vector and generates a representation for each sentence. Finally, the sentence-level attention layer is used to extract sentences that provide the most significant information about the document class. A document vector that contains the summary of the entire sample is generated using these informative sentences. The neural network architecture details are described in \cite{yang2016hierarchical}.   
\end{itemize}

\begin{table}
\begin{center}
\caption{Models and their accuracies on datasets}
\label{tab2}
 \begin{tabular*}{\textwidth}{c @{\extracolsep{\fill}} cccccc}
 \hline
Model & BBC News & AG News & 20NG & BBC Sports & IMDB & R8 \\
        \hline
        TFIDF with & 95.73 & 90.45 & 81.69 & 91.21 & 82.99 & 84.24  \\ Naive-Bayes & & & & & &\\
        \hline
        GloVe+Average & 94.16 & 92.07 & 80.43 & 97.30 & 86.31 & 95.57  \\
        \hline
        GloVe+Attention & 95.28 & 92.39 & 81.65 & 95.95 & 88.76 & 95.61  \\ 
        \hline
        LSTM+CNN & 96.18 & 92.71 & 79.74  & \textbf{
        99.32} & 89.15 & 97.17 \\
        \hline
        BiLSTM+Max & 95.73 & 92.59 & 83.02 & 97.97 & 88.88 & 97.03   \\
        \hline
        BiLSTM+Attention & 96.63 & 93.14 & 81.76 & 98.65 & 89.29 & 95.80  \\
        \hline
        USE & 96.63 & 92.09 & 81.76 & 98.65 & 87.14 & 95.61 \\
        \hline
        ULMFiT & 97.07 & 94.00 & 82.4 & 98.65 & 92.6 & 96.48 \\
        \hline
        HAN & 97.75 & 92.11 & 85.01 & 96.24 & 88.94 & 94.47 \\
        \hline
        BERT & \textbf{98.2} & \textbf{94.04} & \textbf{85.78} & 98.65 & \textbf{95.63} & \textbf{97.62} \\
        \hline
        DistilBERT & 97.3 & 94.02 & 85.43 & \textbf{99.32} & 92.18 & 97.53 \\
        \hline
 \end{tabular*}    
\end{center}

\end{table}

\begin{figure}
 \includegraphics[width=0.45\textwidth]{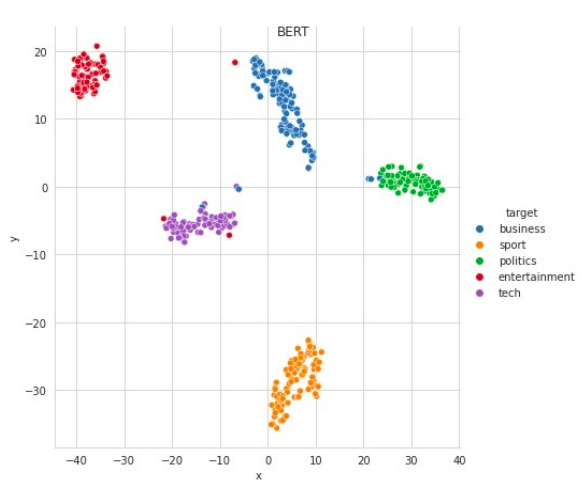}
 \includegraphics[width=0.45\textwidth]{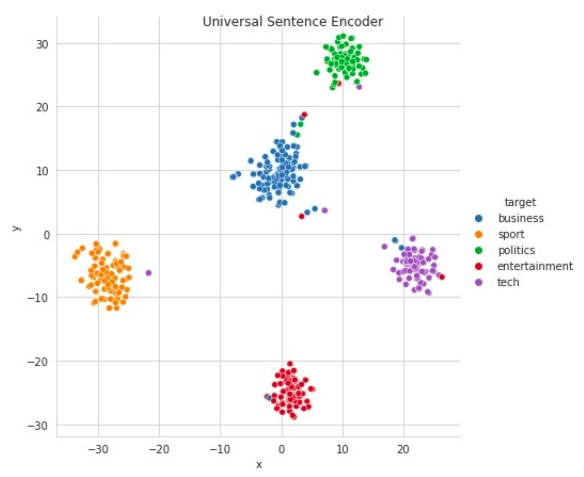}
 \caption{t-SNE Plots of BERT and USE on BBC News Dataset with fine-tuning}
 \label{fig1}
 \end{figure}

 \begin{figure}
 \includegraphics[width=0.45\textwidth]{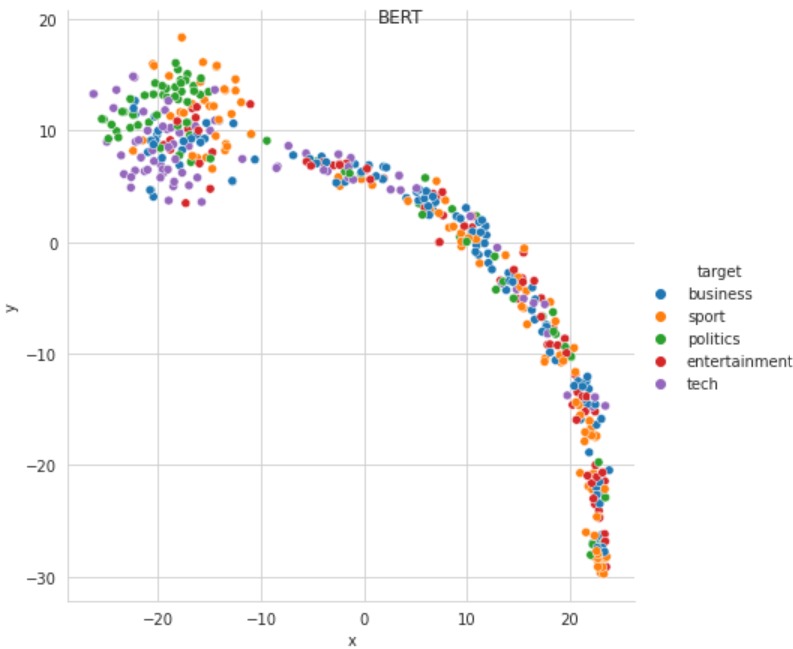}
 \includegraphics[width=0.45\textwidth]{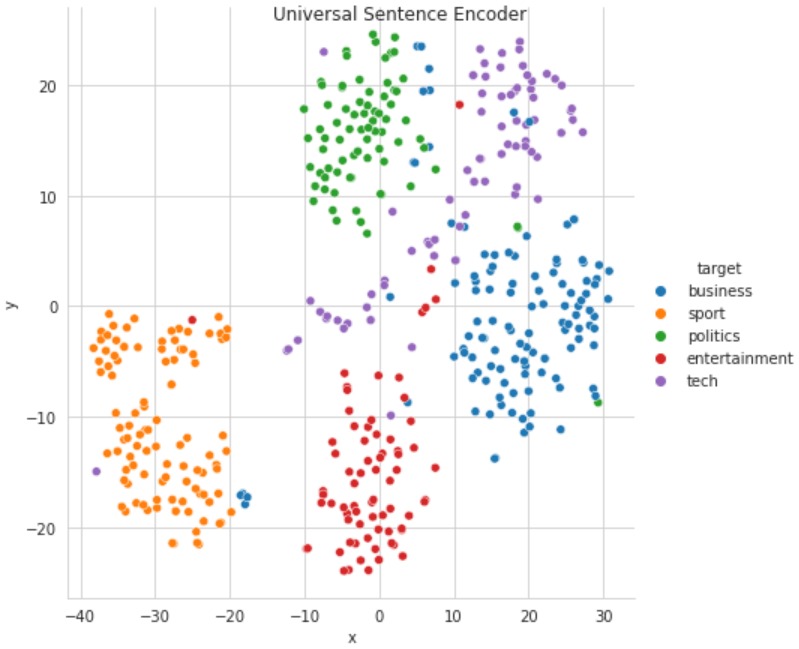}
 \caption{t-SNE Plots of BERT and USE on BBC News Dataset without fine-tuning}
 \label{fig2}
 \end{figure}

\section{Results}
In this work, we perform an extensive comparison of machine learning techniques on six standard text classification datasets. Experiments have been performed with a variety of models with different complexities. Table \ref{tab2} gives an overview of the experimental results of different architectures on the datasets. The BERT and DistilBERT models give the highest accuracy for most of the datasets. In general, the pre-trained models USE, ULMFiT, BERT, and DistilBERT models perform better than raw models trained from scratch. However, even the baseline models like naive Bayes and Glove + Max/Attention provides a reasonable baseline in terms of accuracy. The performance simple Glove + attention is comparable to the best models on most of the datasets. This signifies that document classification is a relatively simpler task and can be accurately done using a bag of words like networks. There is an exception for the IMDB data set where we clearly see the benefit of using pre-trained deep BERT-based models. This shows that sentiment analysis is a difficult task and needs sophisticated models as compared to regular document classification.

The importance of using pre-trained embeddings is seen in BBC Sports, IMDB, and R8 datasets. The Glove + Max/Attention provides a reasonable benefit over TF-IDF + NB on these datasets. In the case of shallow neural models, BiLSTM + Max is a good choice as it provides decent performance across all the datasets. Out of the shallow pre-trained models like ULMFiT and USE, ULMFiT performs better on most of the datasets. The performance of USE is similar to its non-pre-trained counterparts and hence is less desirable. Even HAN gives comparable accuracy without any pre-training involved.

The t-SNE plots used for visualization of BERT and USE embeddings on BBC News dataset with and without fine-tuning the corresponding models are shown in Fig \ref{fig1} and \ref{fig2}. The representations learned by pre-trained USE without finetuning are more distinctive as compared to BERT. However, after finetuning the BERT representations are more separable and the same is reflected in the accuracy metric. 

\section{Conclusion}
We have analyzed a host of selective models with variable complexities on standard document classification datasets.  The models evaluated include basic machine learning models like naive Bayes classifier, sequential models like LSTM + CNN, BiLSTM + Attention, HAN, USE, ULMFiT, and transformer-based models like BERT and DistilBERT. The models are benchmarked on BBC News, AG News, 20 NewsGroup (20 NG), R8, BBC Sports, and IMDB datasets. The pre-trained models show superior performance as compared to raw models. However, the simpler bag of words model like Glove + Attention/Max pooling can be used instead of more complex models with minimal loss in accuracy in most of the datasets. Even the shallow BiLSTM + Max architecture performs decently across all the datasets and can be a reasonable choice when using non-pretrained shallow models. The importance of using sophisticated models is notably seen on the IMDB sentiment classification dataset. Overall we indicate that choice of algorithm is dataset dependent and simpler alternatives can be explored as opposed to the recent trends of BERT only inclination.

\section*{Acknowledgements}
This research was conducted under the guidance of L3Cube, Pune. We would like to express our gratitude towards our mentors at L3Cube for their continuous support and encouragement. 

\bibliographystyle{splncs04}
\bibliography{main.bib}
%




\end{document}